# A Corpus and Cloze Evaluation for Deeper Understanding of Commonsense Stories

**Nasrin Mostafazadeh**[1], **Nathanael Chambers**[2], **Xiaodong He**[3], **Devi Parikh**[4], **Dhruv Batra**[4], **Lucy Vanderwende**[3], **Pushmeet Kohli**[3], **James Allen**[1,5]

1 University of Rochester, 2 United States Naval Academy, 3 Microsoft Research, 4 Virginia Tech, 5 The Institute for Human & Machine Cognition

{nasrinm,james}@cs.rochester.edu, nchamber@usna.edu, {parikh,dbatra}@vt.edu, {xiaohe,lucyv,pkohli}@microsoft.com

## Abstract

Representation and learning of commonsense knowledge is one of the foundational problems in the quest to enable deep language understanding. This issue is particularly challenging for understanding casual and correlational relationships between events. While this topic has received a lot of interest in the NLP community, research has been hindered by the lack of a proper evaluation framework. This paper attempts to address this problem with a new framework for evaluating story understanding and script learning: the 'Story Cloze Test'. This test requires a system to choose the correct ending to a four-sentence story. We created a new corpus of 50k five-sentence commonsense stories, ROCStories, to enable this evaluation. This corpus is unique in two ways: (1) it captures a rich set of causal and temporal commonsense relations between daily events, and (2) it is a high quality collection of everyday life stories that can also be used for story generation. Experimental evaluation shows that a host of baselines and state-of-the-art models based on shallow language understanding struggle to achieve a high score on the Story Cloze Test. We discuss these implications for script and story learning, and offer suggestions for deeper language understanding.

## 1 Introduction

Story understanding is an extremely challenging task in natural language understanding with a long-running history in AI (Charniak, 1972; Winograd, 1972; Turner, 1994; Schubert and Hwang, 2000). Recently, there has been a renewed interest in story and narrative understanding based on progress made in core NLP tasks. This ranges from generic story telling models to building systems which can compose meaningful stories in collaboration with humans (Swanson and Gordon, 2008). Perhaps the biggest challenge of story understanding (and story generation) is having commonsense knowledge for the interpretation of narrative events. The question is how to provide commonsense knowledge regarding daily events to machines.

A large body of work in story understanding has focused on learning scripts (Schank and Abelson, 1977). Scripts represent structured knowledge about stereotypical event sequences together with their participants. It is evident that various NLP applications (text summarization, co-reference resolution, question answering, etc.) can hugely benefit from the rich inferential capabilities that structured knowledge about events can provide. Given that developing hand-built scripts is extremely time-consuming, there is a serious need for automatically induced scripts. Most relevant to this issue is work on unsupervised learning of 'narrative chains' (Chambers and Jurafsky, 2008) and event schemas (Chambers and Jurafsky, 2009; Balasubramanian et al., 2013; Cheung et al., 2013; Nguyen et al., 2015). The first requirement of any learner is to decide on a corpus to drive the learning process. We are foremost interested in a resource that is full of temporal and causal relations between events because causality is a central component of coherency. Personal stories from daily weblogs are good sources of commonsense causal information (Gordon and Swan-son, 2009; Manshadi et al., 2008), but teasing out useful information from noisy blog entries is a problem of its own. Consider the following snippet from ICWSM 2011 Spinn3r Dataset of Weblog entries (Burton et al., 2009):

> "I had an interesting day in the studio today. It was so interesting that I took pictures along the way to describe it to you. Sometimes I like to read an autobiography/biography to discover how someone got from there to here.....how they started, how they traveled in mind and spirit, what made them who they are now. Well, today, my work was a little like that."

This text is full of discourse complexities. A host of challenging language understanding tasks are required to get at the commonsense knowledge embedded within such text snippets. What is needed is a simplified version of these narratives. This paper introduces a new corpus of such short commonsense stories. With careful prompt design and multiple phases of quality control, we collected 50k high quality five-sentence stories that are full of stereotypical causal and temporal relations between events. The corpus not only serves as a resource for learning commonsense narrative schemas, but is also suitable for training story generation models. We describe this corpus in detail in Section 3.

This new corpus also addresses a problem facing script learning over the past few years. Despite the attention scripts have received, progress has been inhibited by the lack of a systematic evaluation framework. A commonly used evaluation is the 'Narrative Cloze Test' (Chambers and Jurafsky, 2008) in which a system predicts a held-out event (a verb and its arguments) given a set of observed events. For example, the following is one such test with a missing event: {X threw, pulled X, told X, ???, X completed}[1]. As is often the case, several works now optimize to this specific test, achieving higher scores with shallow techniques. This is problematic because the models often are not learning commonsense knowledge, but rather how to beat the shallow test.

This paper thus introduces a new evaluation framework called the Story Cloze Test. Instead of predicting an event, the system is tasked with choosing an entire sentence to complete the given story. We collected 3,742 doubly verified Story Cloze Test cases. The test is described in detail in Section 4.

Finally, this paper proposes several models, including the most recent state-of-the-art approaches for the narrative cloze test, for tackling the Story Cloze Test. The results strongly suggest that achieving better than random or constant-choose performance requires richer semantic representation of events together with deeper levels of modeling the semantic space of stories. We believe that switching to the Story Cloze Test as the empirical evaluation framework for story understanding and script learning can help direct the field to a new direction of deeper language understanding.

## 2 Related Work

Several lines of research have recently focused on learning narrative/event representations. Chambers and Jurafsky first proposed narrative chains (Chambers and Jurafsky, 2008) as a partially ordered set of narrative events that share a common actor called the 'protagonist'. A narrative event is a tuple of an event (a verb) and its participants represented as typed dependencies. Several expansions have since been proposed, including narrative schemas (Chambers and Jurafsky, 2009), script sequences (Regneri et al., 2010), and relgrams (Balasubramanian et al., 2013). Formal probabilistic models have also been proposed to learn event schemas and frames (Cheung et al., 2013; Bamman et al., 2013; Chambers, 2013; Nguyen et al., 2015). These are trained on smaller corpora and focus less on large-scale learning. A major shortcoming so far is that these models are mainly trained on news articles. Little knowledge about everyday life events are learned.

Several groups have directly addressed script learning by focusing exclusively on the narrative cloze test. Jans et al. (Jans et al., 2012) redefined the test to be a text ordered sequence of events, whereas the original did not rely on text order (Chambers and Jurafsky, 2008). Since then, others have shown language-modeling techniques perform well (Pichotta and Mooney, 2014a; Rudinger et al., 2015). This paper shows that these approaches struggle on the richer Story Cloze evaluation.

There has also been renewed attention toward natural language comprehension and commonsense reasoning (Levesque, 2011; Roemmele et al., 2011; Bowman et al., 2015). There are a few recent frameworks for evaluating language comprehension (Hermann et al., 2015; Weston et al., 2015), including the MCTest (Richardson et al., 2013) as a notable one. Their framework also involves story comprehension, however, their stories are mostly fictional, on average 212 words, and geared toward children in grades 1-4. Some progress has been made in story understanding by limiting the task to the specific domains and question types. This includes research on understanding newswire involving terrorism scripts (Mueller, 2002), stories about people in a restaurant where a reasonable number of questions about time and space can be answered (Mueller, 2007), and generating stories from fairy tales (McIntyre and Lapata, 2009). Finally, there is a rich body of work on story plot generation and creative or artistic story telling (Méndez et al., 2014; Riedl and León, 2008). This paper is unique to these in its corpus of short, simple stories with a wide variety of commonsense events. We show these to be useful for learning, but also for enabling a rich evaluation framework for narrative understanding.

[1] Narrative cloze tests were not meant to be human solvable.

## 3 A Corpus of Short Commonsense Stories

We aimed to build a corpus with two goals in mind:

1. The corpus contains a *variety* of commonsense causal and temporal relations between everyday events. This enables learning narrative structure across a range of events, as opposed to a single domain or genre.

2. The corpus is a high quality collection of nonfictional daily short life stories, which can be used for training rich coherent story-telling models.

In order to narrow down our focus, we carefully define a narrative or story as follows: 'A narrative or story is anything which is told in the form of a causally (logically) linked set of events involving some shared characters'. The classic definition of a story requires having a plot, (e.g., a character following a goal and facing obstacles), however, here we are not concerned with how entertaining or dramatic the stories are. Instead, we are concerned with the essence of actually being a logically meaningful story. We follow the notion of 'storiness' (Forster, 1927; Bailey, 1999), which is described as "the expectations and questions that a reader may have as the story develops", where expectations are 'common-sense logical inferences' made by the imagined reader of the story.

We propose to satisfy our two goals by asking hundreds of workers on Amazon Mechanical Turk (AMT) to write novel five-sentence stories. The five-sentence length gives enough context to the story without allowing room for sidetracks about less important or irrelevant information in the story. In this Section we describe the details about how we collected this corpus, and provide statistical analysis.

### 3.1 Data Collection Methodology

Crowdsourcing this corpus makes the data collection scalable and adds to the diversity of stories. We tested numerous pilots with varying prompts and instructions. We manually checked the submitted stories in each pilot and counted the number of submissions which did not have our desired level of coherency or were specifically fictional or offensive. Three people participated in this task and they iterated over the ratings until everyone agreed with the next pilot's prompt design. We achieved the best results when we let the workers write about anything they have in mind, as opposed to mandating a prespecified topic. The final crowdsourcing prompt can be found in supplementary material.

The key property that we had enforced in our final prompt was the following: the story should read like a coherent story, with a specific *beginning* and *ending*, where *something happens* in between. This constraint resulted in many causal and temporal links between events. Table 1 shows the examples we provided to the workers for instructing them about the constraints. We set a limit of 70 characters to the length of each sentence. This prevented multi-part sentences that include unnecessary details. The workers were also asked to provide a title that best describes their story. Last but not least, we instructed the workers not to use quotations in their sentences and avoid using slang or informal language.

Collecting high quality stories with these constraints gives us a rich collection of commonsense stories which are full of stereotypical inter-event re-

| | |
|---|---|
| ✗ | The little puppy thought he was a great basketball player. He challenged the kitten to a friendly game. The kitten agreed. Kitten started to practice really hard. Eventually the kitten beat the puppy by 40 points. |
| ✓ | Bill thought he was a great basketball player. He challenged Sam to a friendly game. Sam agreed. Sam started to practice really hard. Eventually Sam beat Bill by 40 points. |
| ✗ | I am happy with my life. I have been kind. I have been successful. I work out. Why not be happy when you can? |
| ✗ | The city is full of people and offers a lot of things to do. One of my favorite things is going to the outdoor concerts. I also like visiting the different restaurants and museums. There is always something exciting to do in the city. |
| ✓ | The Smith family went to the family beach house every summer. They loved the beach house a lot. Unfortunately there was a bad hurricane once. Their beach house was washed away. Now they lament the loss of their beach house every summer. |
| ✗ | Miley was in middle school. ~~She lived in an apartment~~. Once Miley made a mistake and cheated in one of her exams. She tried to hide the truth from her parents. After her parents found out, they grounded her for a month. |
| ✓ | Miley was in middle school. She usually got good grades in school . Once Miley made a mistake and cheated in one of her exams. She tried to hide the truth from her parents. After her parents found out, they grounded her for a month. |

Table 1: Examples of good and bad stories provided to the crowd-sourced workers. Each row emphasizes one of the three properties that each story should satisfy: (1) being realistic, (2) having clear beginning and ending, and (3) not stating anything irrelevant to the story.

X *challenge* Y – Y *agree* play —— Y *practice* —— Y *beat* X

Figure 1: An example narrative chain with characters X and Y.

lations. For example, from the good story in first row of Table 1, one can extract the narrative chain represented in Figure 1. Developing a better semantic representation for narrative chains which can capture rich inter-event relations in these stories is a topic of future work.

**Quality Control:** One issue with crowdsourcing is how to instruct non-expert workers. This task is a type of creative writing, and is trickier than classification and tagging tasks. In order to ensure we get qualified workers, we designed a qualification test on AMT in which the workers had to judge whether or not a given story (total five stories) is an acceptable one. We used five carefully selected stories to be a part of the qualification test. This not only eliminates any potential spammers on AMT, but also provides us with a pool of creative story writers. Furthermore, we qualitatively browsed through the submissions and gave the workers detailed feedback before approving their submissions. We often bonused our top workers, encouraging them to write new stories on a daily basis.

**Statistics:** Figure 2 shows the distribution of number of tokens of different sentence positions. The first sentence tends to be shorter, as it usually introduces characters or sets the scene, and the fifth sentence is longer, providing more detailed conclusions to the story. Table 2 summarizes the statistics of our crowdsourcing effort. Figure 3 shows the distribution of the most frequent 50 events in the corpus. Here we count event as any hyponym of 'event' or 'process' in WordNet (Miller., 1995). The top two events, 'go' and 'get', each comprise less than 2% of all the events, which illustrates the rich diversity of the corpus.

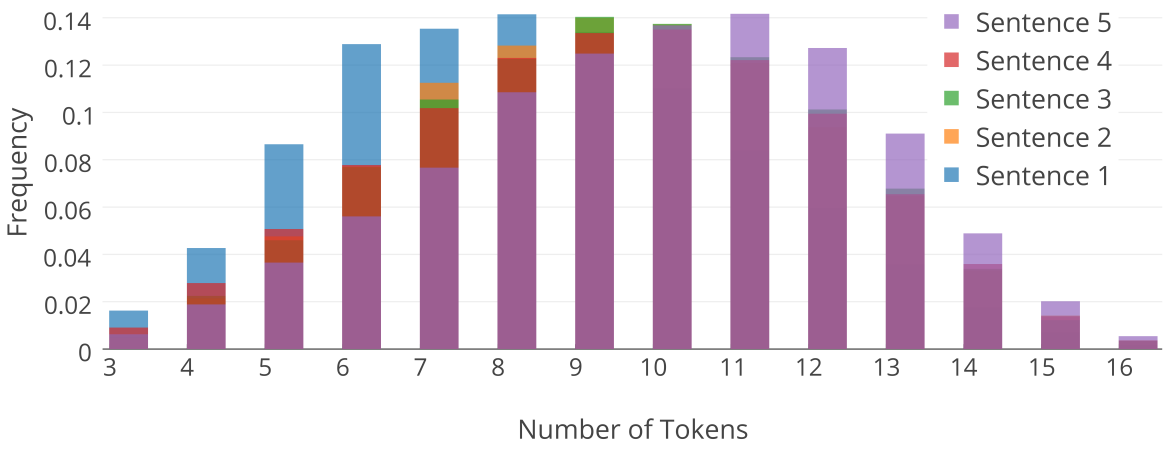

Figure 2: Number of tokens in each sentence position.

| | |
|---|---|
| # submitted stories | 49,895 |
| # approved stories | **49,255** |
| # workers participated | 932 |
| Average # stories by one worker | 52.84 |
| Max # stories written by one worker | 3,057 |
| Average work time among workers (minute) | 4.80 |
| Median work time among workers (minute) | 2.16 |
| Average payment per story (cents) | 26 |

Table 2: Crowdsourcing worker statistics.

Figure 4 visualizes the n-gram distribution of our story titles, where each radial path indicates an n-gram sequence. For this analysis we set n=5, where the mean number of tokens in titles is 9.8 and median is 10. The 'end' token distinguishes the actual ending of a title from five-gram cut-off. This figure demonstrates the range of topics that our workers have written about. The full circle reflects on 100% of the title n-grams and the n-gram paths in the faded 3/4 of the circle comprise less than 0.1% of the n-grams. This further demonstrates that the range of topics covered by our corpus is quite diverse. A full dynamic visualization of these n-grams can be found here: `http://goo.gl/Qhg60B`.

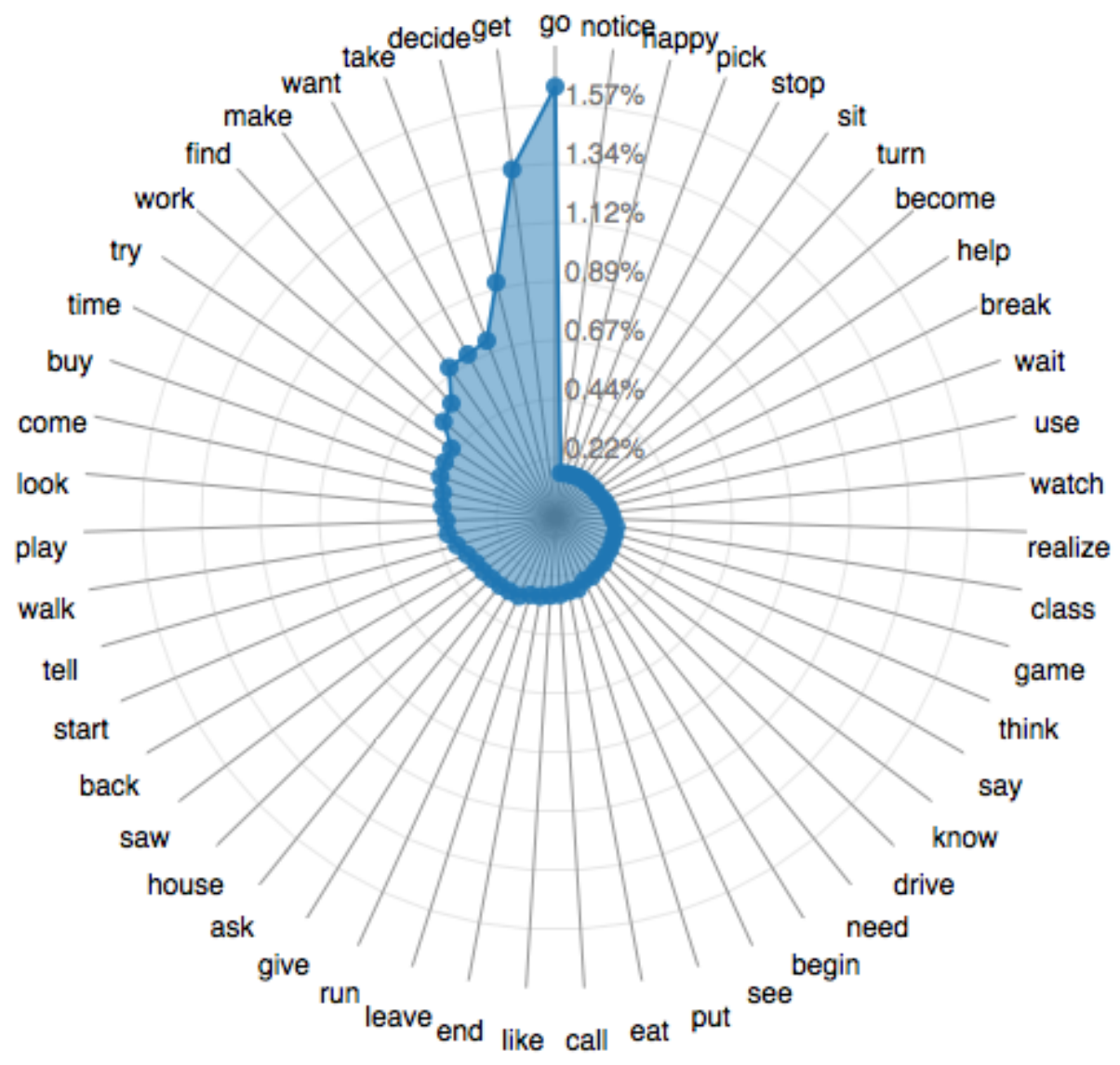

Figure 3: Distribution of top 50 events in our corpus.

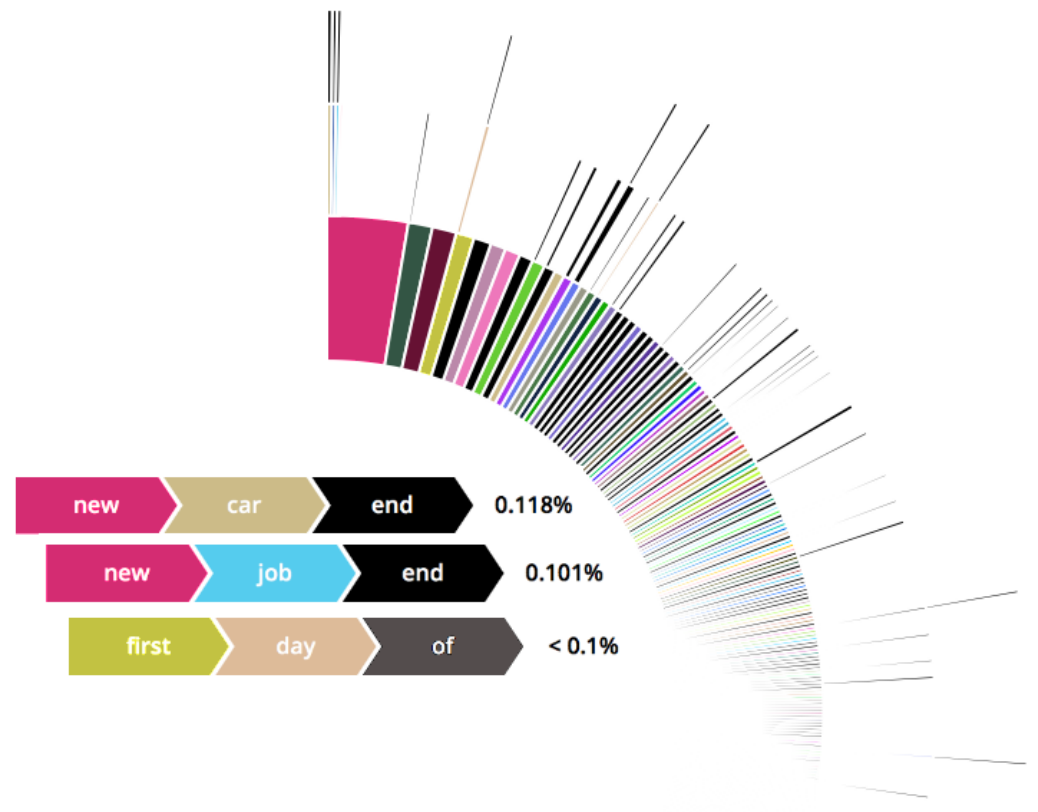

Figure 4: N-gram distribution of story titles.

### 3.2 Corpus Release

The corpus is publicly available to the community and can be accessed through `http://cs.rochester.edu/nlp/rocstories`, which will be grown even further over the coming years. Given the quality control pipeline and the creativity required from workers, data collection goes slowly.

We are also making available semantic parses of these stories. Since these stories are not newswire, off-the-shelf syntactic and shallow semantic parsers for event extraction often fail on the language. To address this issue, we customized search parameters and added a few lexical entries[2] to TRIPS broad-coverage semantic parser[3], optimizing its performance on our corpus. TRIPS parser (Allen et al., 2008) produces state-of-the-art logical forms for input stories, providing sense disambiguated and ontology-typed rich deep structures which enables event extraction together with semantic roles and coreference chains throughout the five sentences.

### 3.3 Temporal Analysis

Being able to temporally order events in the stories is a pre-requisite for complete narrative understanding. Temporal analysis of the events in our short commonsensical stories is an important topic of further research on its own. In this Section, we summarize two of our analyses regarding the nature of temporal ordering of events in our corpus.

**Shuffling Experiment:** An open question in any text genre is how text order is related to temporal order. Do the sentences follow the real-world temporal order of events? This experiment shuffles the stories and asks AMT workers to arrange them back to a coherent story. This can shed light on the correlation between the original position of the sentences and the position when another human rearranges them in a commonsensically meaningful way. We set up this experiment as follows: we sampled two sets of 50 stories from our corpus: *Good-Stories*$_{50}$ and *Random-Stories*$_{50}$. *Good-Stories*$_{50}$[4] is sampled from a set of stories written by top workers who have shown shown consistent quality throughout their submissions. *Random-Stories*$_{50}$[5] is a random sampling from all the stories in the corpus. Then we randomly shuffled the sentences in each story and asked five crowd workers on AMT to rearrange the sentences.

[2]For example, new informal verbs such as 'vape' or 'vlog' have been added to the lexicon of this semantic parser.

[3]`http://trips.ihmc.us/parser/cgi/step`

[4]This set can be found here: `https://goo.gl/VTnJ9s`

| | *Good-Stories*$_{50}$ | *Random-Stories*$_{50}$ |
|---|---|---|
| % perfectly ordered, taking majority ordering for each of the 50 stories | 100 | 86 |
| % all sentences perfectly ordered, out of 250 orderings | 95.2 | 82.4 |
| % $\leq$ 1 sentences misplaced, rest flow correctly, out of 250 orderings | 98.0 | 96.0 |
| % correct placements of each position, 1 to 5 | **98.8**, 97.6, 96, 96, **98.8** | **95.6**, 86, 86.8, 91.2, **96.8** |

Table 3: Results from the human temporal shuffling experiment.

Table 3 summarizes the results of this experiment. The first row shows the result of ordering if we take the absolute majority ordering of the five crowd workers as the final ordering. The second row shows the result of ordering if we consider each of the 250 (50 stories x 5 workers ordering each one) ordering cases independently. As shown, the good stories are perfectly ordered with very high accuracy. It is important to note that this specific set rarely had any linguistic adverbials such as 'first', 'then', etc. to help human infer the ordering, so the main factors at play are the following: (1) the commonsensical temporal and causal relation between events (narrative schemas), e.g., human knows that first someone loses a phone then starts searching; (2) the natural way of narrating a story which starts with introducing the characters and concludes the story at the end. The role of the latter factor is quantified in the misplacement rate of each position reported in Table 3, where the first and last sentences are more often correctly placed than others. The high precision of ordering in sentences 2 up to 4 further verifies the richness of our corpus in terms of logical relation between events.

**TimeML Annotation:** TimeML-driven analysis of these stories can give us finer-grained insight about temporal aspect of the events in this corpus. We performed a simplified TimeML-driven (Pustejovsky et al., 2003) expert annotation of a sample of 20 stories[6]. Among all the temporal links (TLINK) annotated, 62% were 'before' and 10% were 'simultaneous'. We were interested to know if the actual text order mirrors real-world order of events. We found that sentence order matches TimeML order 55% of the time. A more comprehensive study of temporal and causal aspects of these stories requires defining a specific semantic annotation framework which covers not only temporal but also causal relations between commonsense events. This is captured in a recent work on semantic annotation of ROCStories (Mostafazadeh et al., 2016).

## 4 A New Evaluation Framework

As described earlier in the introduction, the common evaluation framework for script learning is the 'Narrative Cloze Test' (Chambers and Jurafsky, 2008), where a system generates a ranked list of guesses for a missing event, given some observed events. The original goal of this test was to provide a comparative measure to evaluate narrative knowledge. However, gradually, the community started optimizing towards the performance on the test itself, achieving higher scores without demonstrating narrative knowledge learning. For instance, generating the ranked list according to the event's corpus frequency (e.g., always predicting 'X said') was shown to be an extremely strong baseline (Pichotta and Mooney, 2014b). Originally, narrative cloze test chains were extracted by hand and verified as gold chains. However, the cloze test chains used in all of the most recent works are not human verified as gold.

It is evident that there is a need for a more systematic automatic evaluation framework which is more in line with the original deeper script/story understanding goals. It is important to note that reordering of temporally shuffled stories (Section 3.3) can serve as a framework to evaluate a system's story understanding. However, reordering can be achieved to a degree by using various surface features such as adverbials, so this cannot be a foolproof story understanding evaluation framework. Our ROCStories corpus enables a brand new framework for evaluating story understanding, called the *'Story Cloze Test'*.

[5]This set can be found here: `https://goo.gl/pgm2KR`

[6]The annotation is available: `http://goo.gl/7qdNsb`

### 4.1 Story Cloze Test

The cloze task (Taylor, 1953) is used to evaluate a human (or a system) for language understanding by deleting a random word from a sentence and having a human fill in the blank. We introduce 'Story Cloze Test', in which a system is given a four-sentence 'context' and two alternative endings to the story, called 'right ending' and 'wrong ending'. Hence, in this test the fifth sentence is blank. Then the system's task is to choose the right ending. The 'right ending' can be viewed as 'entailing' hypothesis in a classic Recognizing Textual Entailment (RTE) framework (Giampiccolo et al., 2007), and 'wrong' ending can be seen as the 'contradicting' hypothesis. Table 4 shows three example Story Cloze Test cases.

Story Cloze Test will serve as a generic story understanding evaluation framework, also applicable to evaluation of story generation models (for instance by computing the log-likelihoods assigned to the two ending alternatives by the story generation model), which does not necessarily imply requirement for explicit narrative knowledge learning. However, it is safe to say that any model that performs well on Story Cloze Test is demonstrating some level of deeper story understanding.

### 4.2 Data Collection Methodology

We randomly sampled 13,500 stories from ROCStories Corpus and presented only the first four sentences of each to AMT workers. For each story, a worker was asked to write a 'right ending' and a 'wrong ending'. The workers were prompted to satisfy two conditions: (1) the sentence should follow up the story by sharing at least one of the characters of the story, and (2) the sentence should be entirely realistic and sensible when read in isolation. These conditions make sure that the Story Cloze Test cases are not trivial. More details on this setup is described in the supplementary material.

**Quality Control:** The accuracy of the Story Cloze Test can play a crucial role in directing the research community in the right trajectory. We implemented the following two-step quality control:

1. Qualification Test: We designed a qualification test for this task, where the workers had to choose whether or not a given 'right ending' and 'wrong ending' satisfy our constraints. At this stage we collected 13,500 cloze test cases.

2. Human Verification: In order to further validate the cloze test cases, we compiled the 13,500 Story Cloze Test cases into $2 \times 13,500 = 27,000$ full five-sentence stories. Then for each story we asked three crowd workers to verify whether or not the given sequence of five sentences makes sense as a meaningful and coherent story, rating within {-1, 0, 1}. Then we filtered cloze test cases which had 'right ending' with all ratings 1 and 'wrong ending' with all ratings 0. This process ensures that there are no boundary cases of 'right ending' and 'wrong ending'. This resulted in final 3,742 test cases, which was randomly divided into validation and test Story Cloze Test sets. We also made sure to remove the original stories used in the validation and test set from our ROCStories Corpus.

**Statistics:** Table 5 summarizes the statistics of our crowdsourcing effort. The Story Cloze Test sets can also be accessed through our website.

## 5 Story Cloze Test Models

In this Section we demonstrate that Story Cloze Test cannot be easily tackled by using shallow techniques, without actually understanding the underlying narrative. Following other natural language inference frameworks such as RTE, we evaluate system performance according to basic accuracy measure, which is defined as $\frac{\#correct}{\#test\ cases}$. We present the following baselines and models for tackling Story Cloze Test. All of the models are tested on the validation and test Story Cloze sets, where only the validation set could be used for any tuning purposes.

**1. Frequency**: Ideally, the Story Cloze Test cases should not be answerable without the context. For example, if for some context the two alternatives are 'He was mad after he won'[7] and 'He was cheerful after he won', the first alternative is simply less probable in real world than the other one. This baseline chooses the alternative with higher search engine[8] hits of the main event (verb) together with its semantic roles (e.g., 'I*poison*flowers' vs 'I*nourish*flowers'). We extract the main verb and its corresponding roles using TRIPS semantic parser.

[7] Given our prompt that the 'wrong ending' sentences should make sense in isolation, such cases should be rare in our dataset.

[8] https://developers.google.com/custom-search/

| Context | Right Ending | Wrong Ending |
|---|---|---|
| Tom and Sheryl have been together for two years. One day, they went to a carnival together. He won her several stuffed bears, and bought her funnel cakes. When they reached the Ferris wheel, he got down on one knee. | Tom asked Sheryl to marry him. | He wiped mud off of his boot. |
| Karen was assigned a roommate her first year of college. Her roommate asked her to go to a nearby city for a concert. Karen agreed happily. The show was absolutely exhilarating. | Karen became good friends with her roommate. | Karen hated her roommate. |
| Jim got his first credit card in college. He didn't have a job so he bought everything on his card. After he graduated he amounted a $10,000 debt. Jim realized that he was foolish to spend so much money. | Jim decided to devise a plan for repayment. | Jim decided to open another credit card. |

Table 4: Three example Story Cloze Test cases, completed by our crowd workers.

| | |
|---|---|
| # cases collected | 13,500 |
| # workers participated | 282 |
| Average # cases written by one worker | 47.8 |
| Max # cases written by one worker | 1461 |
| Average payment per test case (cents) | 10 |
| Size of the final set (verified by human) | **3,744** |

Table 5: Statistics for crowd-sourcing Story Cloze Test instances.

**2. N-gram Overlap**: Simply chooses the alternative which shares more n-grams with the context. We compute Smoothed-BLEU (Lin and Och, 2004) score for measuring up to four-gram overlap of an alternative and the context.

**3. GenSim: Average Word2Vec**: Choose the hypothesis with closer average word2vec (Mikolov et al., 2013) embedding to the average word2vec embedding of the context. This is basically an enhanced word overlap baseline, which accounts for semantic similarity.

**4. Sentiment-Full**: Choose the hypothesis that matches the average sentiment of the context. We use the state-of-the-art sentiment analysis model (Manning et al., 2014) which assigns a numerical value from 1 to 5 to a sentence.

**5. Sentiment-Last**: Choose the hypothesis that matches the sentiment of the last context sentence.

**6. Skip-thoughts Model**: This model uses Skip-thoughts' Sentence2Vec embedding (Kiros et al., 2015) which models the semantic space of novels. This model is trained on the 'BookCorpus' (Zhu et al., 2015) (containing 16 different genres) of over 11,000 books. We use the skip-thoughts embedding of the alternatives and contexts for making decision the same way as with GenSim model.

**7. Narrative Chains-AP**: Implements the standard approach to learning chains of narrative events based on Chambers and Jurafsky (2008). An event is represented as a verb and a typed dependency (e.g., the *subject* of *runs*). We computed the PMI between all event pairs in the Associate Press (AP) portion of the English Gigaword Corpus that occur at least 2 times. We run coreference over the given story, and choose the hypothesis whose coreferring entity has the highest average PMI score with the entity's chain in the story. If no entity corefers in both hypotheses, it randomly chooses one of the hypotheses.

**8. Narrative Chains-Stories**: The same model as above, but trained on ROCStories.

**9. Deep Structured Semantic Model (DSSM)**: This model (Huang et al., 2013) is trained to project the four-sentences context and the fifth sentence into the same vector space. It consists of two separate deep neural networks for learning jointly the embedding of the four-sentences context and the fifth sentence, respectively. As suggested in Huang et al. (2013), the input of the DSSM is based on context-dependent characters, e.g., the distribution count of letter-trigrams in the context and in the fifth sentence, respectively. The hyper parameters of the DSSM is determined on the validation set, while the model's parameters are trained on the ROCStories corpus. In our experiment, each of the two neural networks in the DSSM has two layers: the dimen-sion of the hidden layer is 1000, and the dimension of the embedding vector is 300. At runtime, this model picks the candidate with the largest cosine similarity between its vector representation and the context's vector representation.

| | Constant-choose-first | Frequency | N-gram-overlap | GenSim | Sentiment-Full | Sentiment-Last | Skip-thoughts | Narrative-Chains-AP | Narrative-Chains-Stories | DSSM | Human |
|---|---|---|---|---|---|---|---|---|---|---|---|
| Validation Set | 0.514 | 0.506 | 0.477 | 0.545 | 0.489 | 0.514 | 0.536 | 0.472 | 0.510 | 0.604 | 1.0 |
| **Test Set** | 0.513 | 0.520 | 0.494 | 0.539 | 0.492 | 0.522 | 0.552 | 0.478 | 0.494 | **0.585** | 1.0 |

Table 6: The accuracy of various models on The Story Cloze validation and test sets.

The results of evaluating these models on the Story Cloze validation and test sets are shown in Table 6. The constant-choose-first (51%) and human performance (100%) is also provided for comparison. Note that these sets were doubly verified by human, hence it does not have any boundary cases, resulting in 100% human performance. The DSSM model achieves the highest accuracy, but only 7.2 points higher than constant-choose-first. Error analysis on the narrative chains model shows why this and other event-based language models are not sufficient for the task: often, the final sentences of our stories contain complex events beyond the main verb, such as 'Bill was highly unprepared' or 'He had to go to a homeless shelter'. Event language models only look at the verb and syntactic relation like 'was-object' and 'go-to'. In that sense, going to a homeless shelter is the same as going to the beach. This suggests the requirement of having richer semantic representation for events in narratives. Our proposed Story Cloze Test offers a new challenge to the community.

## 6 Discussion

There are three core contributions in this paper: (1) a new corpus of commonsense stories, called ROCStories, (2) a new evaluation framework to evaluate script/story learners, called Story Cloze Test, and (3) a host of first approaches to tackle this new test framework. ROCStories Corpus is the first crowdsourced corpus of its kind for the community. We have released about 50k stories, as well as validation and test sets for Story Cloze Test. This dataset will eventually grow to 100k stories, which will be released through our website. In order to continue making meaningful progress on this task, although it is possible to keep increasing the size of the training data, we expect the community to develop models that will learn to generalize to unseen commonsense concepts and situations.

The Story Cloze Test proved to be a challenge to all of the models we tested. We believe it will serve as an effective evaluation for both story understanding and script knowledge learners. We encourage the community to benchmark their progress by reporting their results on Story Cloze test set. Compared to the previous Narrative Cloze Test, we found that one of the early models for that task actually performs worse than random guessing. We can conclude that Narrative Cloze test spurred interest in script learning, however, it ultimately does not evaluate deeper knowledge and language understanding.

## Acknowledgments

We would like to thank the amazing crowd workers whose endless hours of daily story writing made this research possible. We thank William de Beaumont and Choh Man Teng for their work on TRIPS parser. We thank Alyson Grealish for her great help in the quality control of our corpus. This work was supported in part by Grant W911NF-15-1-0542 with the US Defense Advanced Research Projects Agency (DARPA), the Army Research Office (ARO) and the Office of Naval Research (ONR). Our data collection effort was sponsored by Nuance Foundation.